\documentclass[review,sort&compress,3p,12pt]{elsarticle}

\usepackage{mathtools,amssymb,xfrac,float,color,soul}
\usepackage{graphicx,caption,subcaption}
\usepackage{natbib}
\usepackage[bookmarks=false]{hyperref}
\DeclareGraphicsExtensions {.pdf, .png , .jpg}
\usepackage[bordercolor=white,backgroundcolor=gray!30,linecolor=black,colorinlistoftodos]{todonotes}

\usepackage{lineno}

\begin{document}

\begin{frontmatter}



\title{Trajectory-Based Recognition of Dynamic Persian Sign Language Using Hidden Markov Model}

% authors:

\author[]{Saeideh Ghanbari Azar%
}
\ead{sghanbariazar@tabrizu.ac.ir}

\author[]{Hadi Seyedarabi\corref{cor1}}
\ead{seyedarabi@tabrizu.ac.ir}

\cortext[cor1]{Corresponding author}

% address:
\address{Faculty of Electrical and Computer Engineering, University of Tabriz, Tabriz, Iran}

\begin{abstract}
Sign Language Recognition (SLR) is an important step in facilitating the communication among deaf people and the rest of society. Existing Persian sign language recognition systems are mainly restricted to static signs which are not very useful in everyday communications. In this study, a dynamic Persian sign language recognition system is presented. A collection of 1200 videos were captured from 12 individuals performing 20 dynamic signs with a simple white glove. The trajectory of the hands, along with hand shape information were extracted from each video using a simple region-growing technique. These time-varying trajectories were then modeled using Hidden Markov Model (HMM) with Gaussian probability density functions as observations. The performance of the system was evaluated in different experimental strategies. Signer-independent and signer-dependent experiments were performed on the proposed system and the average accuracy of 97.48\% was obtained. The experimental results demonstrated that the performance of the system is independent of the subject and it can also perform excellently even with a limited number of training data.
\end{abstract}


\end{frontmatter}

%\linenumbers

\section{Introduction}
\label{section.Intro}
Sign language consists of a set of manual or non-manual gestures which are used for communication especially among deaf people. The majority of the gestures of a sign language are manual gestures, while non-manual gestures like head movements (e.g. nodding), body movement (e.g. shrugging) and face expressions also play an important role in sign communications. Unfortunately, the use of sign language is usually restricted to the deaf community resulting in restricted communication of them with the rest of the world. Therefore, they are usually excluded from society and deprived of their rights to have equal educational and career opportunities. In order to address this problem, Sign Language Recognition (SLR) systems are developed to translate this language into speech or text. Developing efficient SLR systems can facilitate the communication of deaf people in society and remove the barriers for them.

One of the basic issues regarding SLR is that there is not a universal sign language. Sign languages of different countries have their own grammar rules. SLR systems have been rapidly developed in recent years for different sign languages including American \cite{vogler1999parallel,wu2015real,lahoti2018android}, Chinese \cite{huang2018novel}, Australian \cite{holden2005australian}, Arabic \cite{tubaiz2015glove,shanableh2007spatio}, Indian \cite{hore2017indian}, Spanish \cite{lopez2012spanish} and Japanese \cite{barricelli2017visual}. For more reviews on sign language and different approaches developed for SLR systems refer to \cite{cheok2019review,parton2005sign}.

Due to its broad range of capabilities, Machine Vision (MV) is the major tool used in the development of SLR systems. An MV-based SLR system usually consists of three components: hand tracker, feature extractor and classifier. Hand tracker’s job is to segment hand regions from the background of the input video frames. Some studies rely on data gloves to track the hand movements for hand tracking \cite{tubaiz2015glove,fang2007large,gao2000sign,khelil2016hand,kumar2017coupled}. Although these gloves are easy and precise to track, they usually contain heavy electromechanical devices which are inconvenient for the signers and limit their natural movements. Another type of studies relies on vision-based methods for hand tracking \cite{shanableh2007spatio,chen2003hand,al2009video}. These techniques usually require some limitations on the signer’s cloths or imaging conditions. For instance, some of the vision-based studies need that the subject wear color gloves to facilitate the hand tracking process \cite{maraqa2008recognition,mohandes2005image,mohandes2012signer}. Nevertheless, these techniques are more convenient and cheaper. Feature extractor is the second stage of an SLR system. It takes the hand tracker’s data and produces a feature vector. Some SLR studies rely on hand shape information to extract a feature vector \cite{mohandes2012signer} while others rely on hand trajectory information \cite{lim2016feature}. Once feature vectors are extracted, they need to be classified using an appropriate classifier. Many different classifiers have been utilized for recognizing different sign languages. These classifiers mainly include Neural Network (NN), K-Nearest Neighbor (KNN) and Hidden Markov Models (HMMs).

Developing machine vision based SLR systems started by the pioneering work of Starner et al. \cite{starner1995visual} in which they developed an American sign language recognition system. They placed the camera on top of a desk or a cap worn by the signer. They used two colored gloves to facilitate the hand tracking stage and classified the signs using HMM. In a recent study, Holden et al. \cite{holden2005australian} presented an HMM-based system which relies on hand shape information to extract a feature vector. This shape information includes hand size, direction, roundedness and the angle between 2 hands. Their system recognizes Australian sign language with the accuracy rate of 97\% at the sentence level and 99\% at the word level. Recently, many researches have been developed regarding Arabic sign language recognition. Al-Rousan et al.\cite{al2009video} suggested a vision-based system that uses Discrete Cosine Transform (DCT) and HMM for recognition of 30 Arabic signs in both offline and online modes. Recognition rates of 96.74\% and 93.8\% were obtained in offline and online mode, respectively. Tubaiz et al. \cite{tubaiz2015glove} proposed a glove-based continuous SLR system. They used a feature extractor which emphasizes the temporal dependency of the data. A modified KNN approach is used for classification. The system recognizes 40 sentences of Arabic sign language with an accuracy of 98.9\%.

Recently, deep learning-based approaches have proven to be very popular in many areas of machine vision applications including sign language recognition. The popularity of these approaches stems from their excellent discriminative abilities and successful performance. The seminal series of studies by Koller et all. \cite{koller2015deep,koller2016deep1,koller2016deep,koller2018deep,koller2019weakly} are among the first studies conducted for deep learning-based sign language recognition. In their early work \cite{koller2015deep}, they have used deep learning for sign language recognition based on the shape of the mouth. In \cite{koller2018deep} they have developed an SLR system which uses Convolutional Neural Networks (CNNs) in an HMM framework. By combining the discriminative abilities of CNNs with the dynamic modeling ability of HMM, they have significantly improved the recognition performance on three benchmark sign language datasets, namely PHOENIX 2012 \cite{koller2015continuous}, PHOENIX 2014 \cite{koller2015continuous}, and SIGNUM \cite{4813472}. In their recent work \cite{koller2019weakly}, they have combined their previous works by adding multi-stream HMM to jointly solve the two sub-problems of hand gesture and mouth shape recognition. They develop a powerful and deep CNN with two bidirectional Long Short-Term Memory (LSTM) layers for recognition of continuous sign language sequences with weak and noisy labels.

\subsection{Related Works to Persian Sign Language (PSL) Recognition}
\label{subsection.Intropersian}
This part focuses on previous attempts made in recognition of PSL. Similar to other sign languages, PSL signs are divided into two main categories, i.e., static signs and dynamic signs. Static signs do not include hand movements and can be captured in a single image. Dynamic signs, on the other hand, include hand movements making it difficult to manipulate. Dynamic signs are usually captured in video frames, and video processing techniques are implemented to recognize them.

Development of a PSL recognition system is in its early stages. There have been few studies conducted in this field. These studies were all focused on image-based recognition of static signs. In the first PSL recognition system, Karami et al. \cite{karami2011persian} collected an image dataset of static alphabet signs. The images were then transformed into the wavelet domain. Different levels of wavelet transform including approximation coefficient of level 6, diagonal and horizontal details of level 6 and 7 and the vertical details of level 6 were used as feature vector. Multi-Layer Perceptron (MLP) neural network was used as a recognizer, and an accuracy rate of 94.06\% was achieved. In another study, Barkoky and Charkari \cite{barkoky2011static} designed a system for recognition of Persian sign numbers. They used a color-based technique to extract hand regions. A thinning method was then applied to these segmented images. The recognition was done by counting the number of endpoints of the thinned image. The accuracy rate of 96.6\% was reported. In a similar study to \cite{karami2011persian}, Moghaddam et al. \cite{moghaddam2011static} reported an image-based system to recognize alphabets of PSL. They used kernel based feature extraction methods including Kernel Principle Component Analysis (KPCA) and Kernel Discriminant Analysis (KDA). Three different classifiers including Minimum Distance (MD), Support Vector Machine (SVM) and NN were used to compare the results. In a more recent study, Zare et al. \cite{zare2018recognition} proposed a recognition system for 10 Persian static signs including six numbers and four words. They used skin segmentation in different color spaces to detect the hand regions. They employed Fourier descriptors as features to train a classifier. Their approach performs good results for real time signer-independent recognition system.

Each of the works discussed above for PSL recognition has its advantages. Since PSL recognition is a newly evolving field of study, there remains some challenges which motivates this paper. All the introduced PSL recognition systems are developed for alphabet or number signs which are not very useful in everyday conversations of the deaf community. This problem indicates the need for developing a dynamic sign recognition system which can recognize more practical signs. Motivated by this, we present a dynamic PSL recognition system. Over 1200 videos of 20 dynamic signs are collected for this system. A region growing technique is used to extract the motion trajectory of the hand and three other shape information. HMM with Gaussian observations is finally utilized to classify 20 dynamic signs.

To summarize, we make the following contributions: First, a new dynamic PSL dataset with 1200 videos is collected which contains 20 single-handed signs that are practical in everyday communication of the deaf community. To increase the diversity of the dataset, 12 individuals are participated in this dataset making it more subject-independent. Second, a dynamic PSL recognition system is proposed which uses a simple trajectory extraction approach based on region growing. The system performs excellently independent of the subject performing the signs, and it has the accuracy of more than 95\% even with a limited number of training data.

The rest of the paper is organized as follows. \hyperref[section.dataset]{Section \ref*{section.dataset}} describes the collected dataset. \hyperref[section.trajectory]{Section \ref*{section.trajectory}} and \hyperref[section.HMM]{Section \ref*{section.HMM}} elaborate the trajectory extraction approach and the HMM classifier, respectively. \hyperref[section.result]{Section \ref*{section.result}}, presents the experiments and compares the results. Finally, the paper is concluded in \hyperref[section.conclusion]{Section \ref*{section.conclusion}}.
\section{Dataset Collection}
\label{section.dataset}
\begin{table}[t]
  \centering
  \caption{List of the signs of the dataset.}
  \label{table.list}
  \resizebox{0.4\textwidth}{0.1\textheight}{%
\begin{tabular}{c c|c c}
  \hline
  Sign Code & Sign & Sign Code & Sign \\
  \hline \hline
  1 & Sad & 11 & Eat \\
  2 & Wish & 12 & Sun \\
  3 & Dear & 13 & Mother \\
  4 & Sorry & 14 & People \\
  5 & How? & 15 & Go \\
  6 & Student & 16 & Day \\
  7 & Today & 17 & Hear \\
  8 & Forget & 18 & Brave \\
  9 & Please & 19 & Natural \\
  10 & Danger & 20 & Can \\
  \hline
\end{tabular}
}
\end{table}
Since there is no dynamic PSL dataset available, the authors needed to construct a dataset. The dataset was collected in the Society of Deaf People (SDP), Urmia, Iran and named as University of Tabriz Persian Sign Language dataset (UoT-PSL) \cite{Azar2016}.In order to develop an efficient recognition system, the collected dataset needs to contain different versions of performing a sign. This can decrease the dependency of the system on the subject. For this purpose, twelve volunteers including six male and six female singers were participated in this dataset. The signers included both deaf and hearing individuals. Twenty dynamic signs of PSL were chosen to be included in the dataset. The signs were selected from the single-handed signs appearing in everyday conversations with the consultation of the experts in SDP. A list of the signs available in this dataset is presented in Table 1. Each individual performed a sign 5 times producing 60 samples for each sign. In the following parts of this paper, the signs will be referred to by their corresponding code in \hyperref[table.list]{Table \ref*{table.list}}.
\begin{figure}
   \centering
   \includegraphics[width=\textwidth]{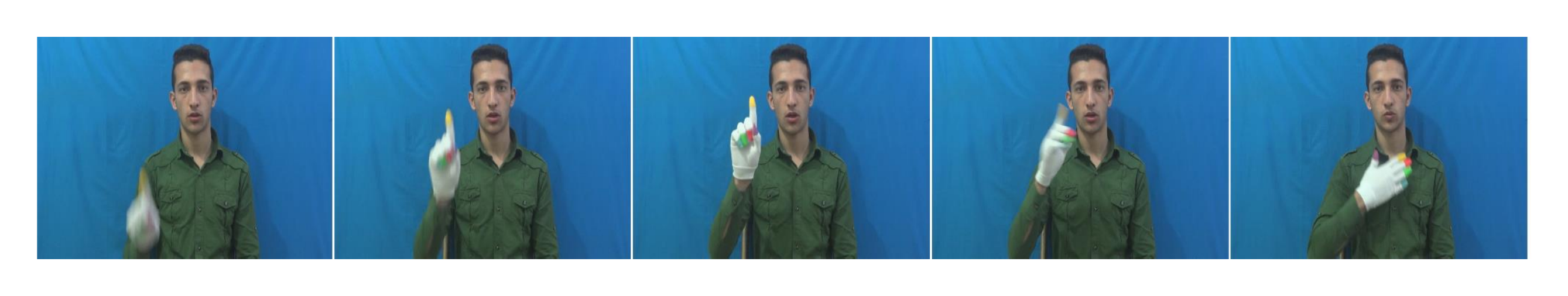}
   \caption{Samples of the captured video frames for sign ‘wish’.}
   \label{fig.behzad}
\end{figure}

A Sony digital camera (model DSC-HX9V) was used to obtain a total number of 1200 videos from 20 signs. There were no particular restrictions on the light of the environment. The collected videos were in AVI format and the frame rate was set to 25 frames per second and the spatial resolution was $1800\times2000$ pixels. The audio contents of the videos were eliminated to avoid unnecessary complexities.

To ease the process of hand tracking, some restrictions were imposed on the imaging conditions. The signers stood in front of a blue background and were asked to wear a simple white glove. Colored marks were considered on the fingertips of the glove for possible future studies but they were not utilized in the present study. All the signs were chosen from single-handed Persian signs to avoid possible occlusions of the hands. \hyperref[fig.behzad]{Figure \ref*{fig.behzad}} presents a sample of the captured video frames.
\section{Sign Trajectory Extraction}
\label{section.trajectory}
\begin{figure}
   \centering
   \includegraphics[width=0.7\textwidth]{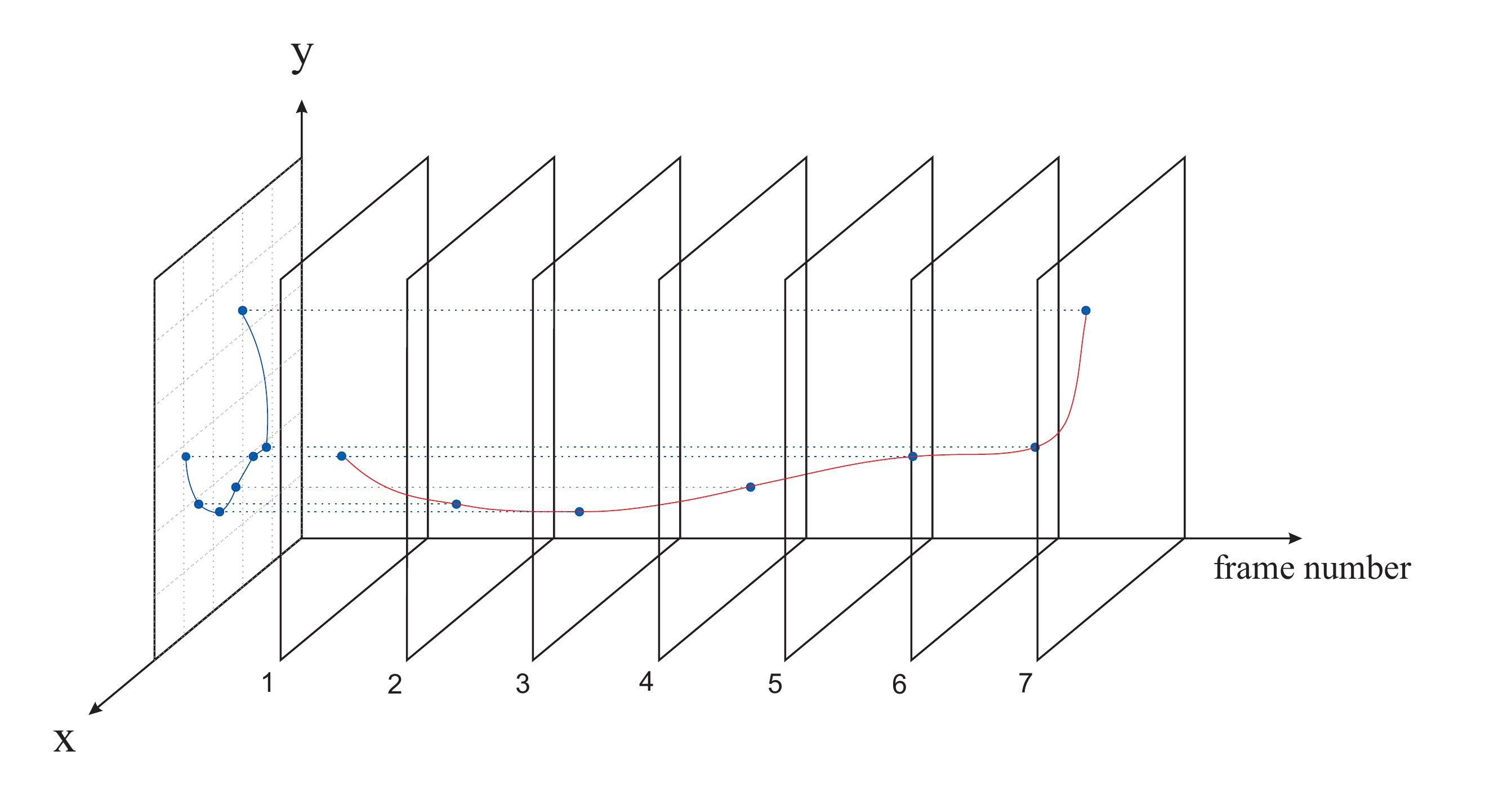}
   \caption{A representation of the hand trajectory.}
   \label{fig.two}
\end{figure}
The first step in developing the PSL recognition system is hand trajectory extraction. That is, in each frame, the hand region is detected, and its centroid in x and y-axes are saved. For a video stream these extracted centroids form the hand trajectory. \hyperref[fig.two]{Figure \ref*{fig.two}} gives an illustration of this definition. The hand region extraction procedure is explained in the following.

\begin{figure}
   \centering
   \includegraphics[width=\textwidth]{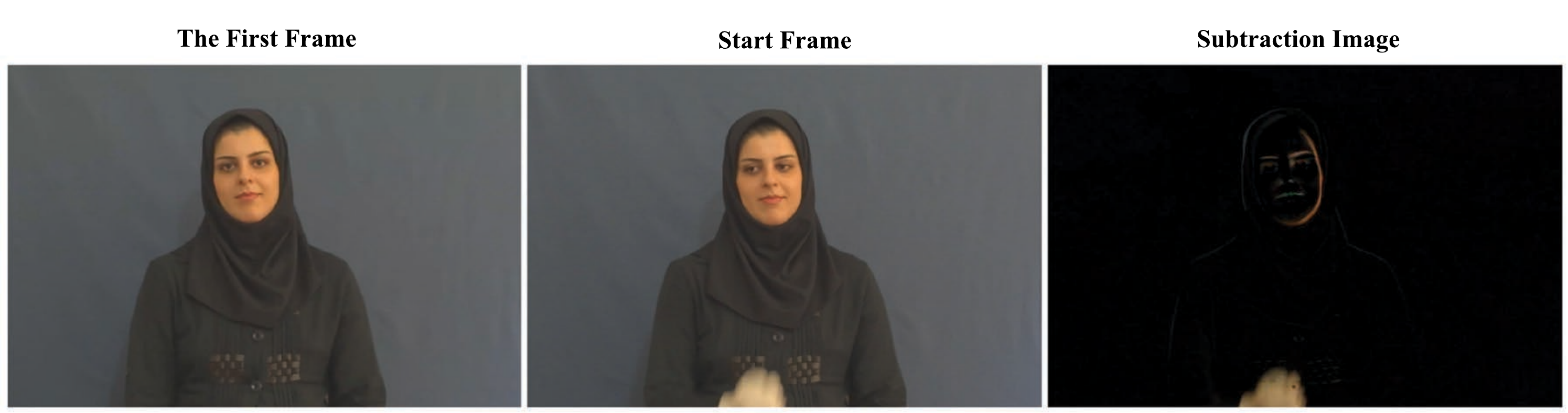}
   \caption{An example of the first frame of the video, start frame and their subtraction image}
   \label{fig.three}
\end{figure}

It should be noted that 10-15 starting frames of the videos did not contain the hand of the signer. In this paper the frame in which the hand appears for the first time in the scene is referred to as \emph{start frame}. Thus, the first step was to detect this so-called start frame. There is no hand region in the frames that come before the start frame.Therefore, they all contain the same image of the signer. This means that subtracting these frames from the first frame results in an approximately zero image. This fact was exploited to detect the start frame. In this regard, each frame was subtracted from the first frame and among this subtracted stream of frames the first one containing a nonzero region bigger than a threshold determined the start frame. \hyperref[fig.three]{Figure \ref*{fig.three}} presents an example of the first frame of the video, start frame and their subtraction image.

\begin{figure}
   \centering
   \includegraphics[width=\textwidth]{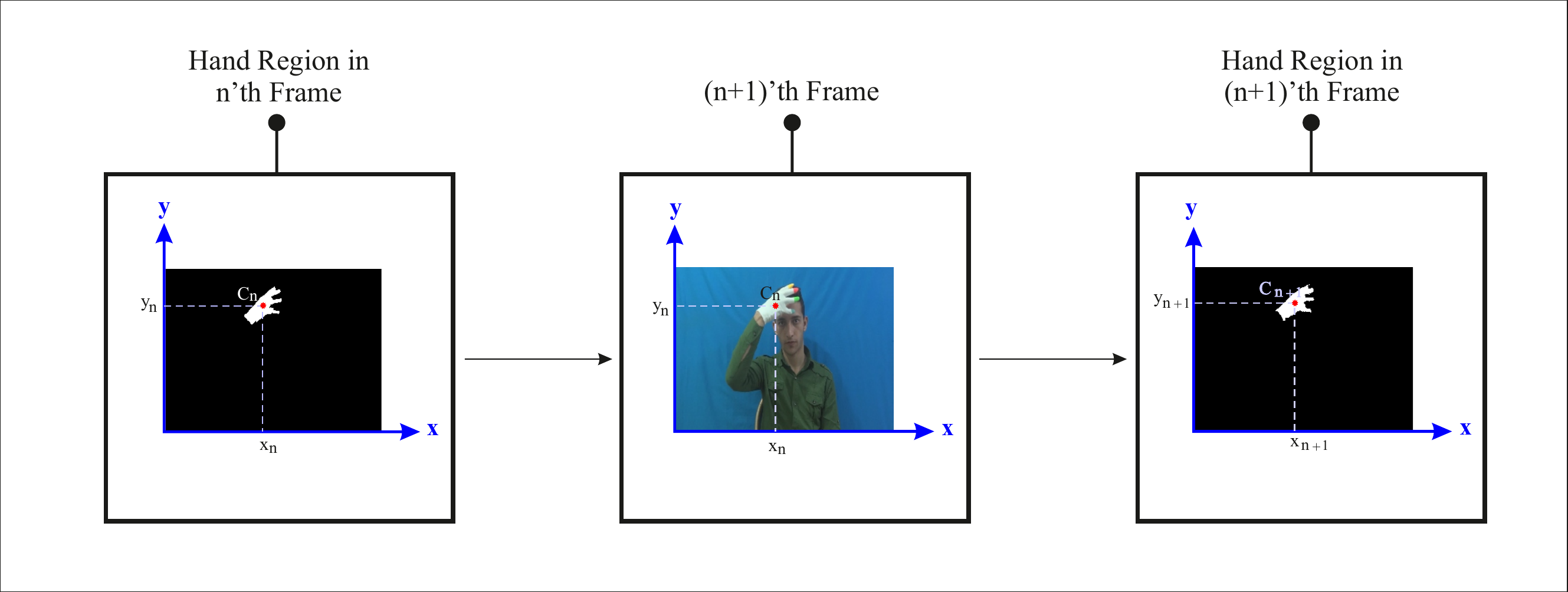}
   \caption{Hand region extraction of $(n+1)$’th frame using the hand centroid of the $n$’th frame as seed for region growing.}
   \label{fig.four}
\end{figure}
\begin{figure}
\centering
    \begin{subfigure}{0.45\textwidth}
         \centering
         \includegraphics[width=\textwidth]{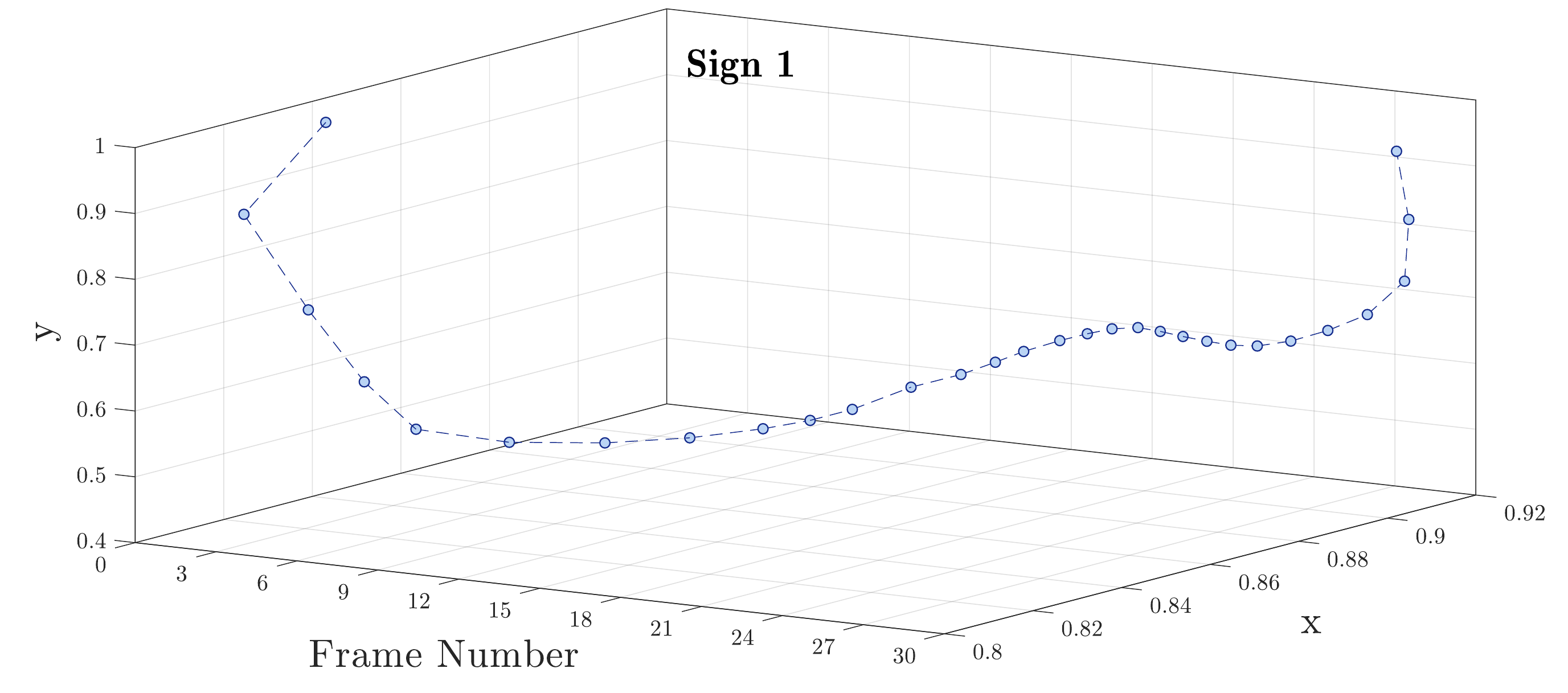}
         \caption{}
        \label{fig.fivea}
    \end{subfigure}
    \begin{subfigure}{0.45\textwidth}
         \centering
         \includegraphics[width=\textwidth]{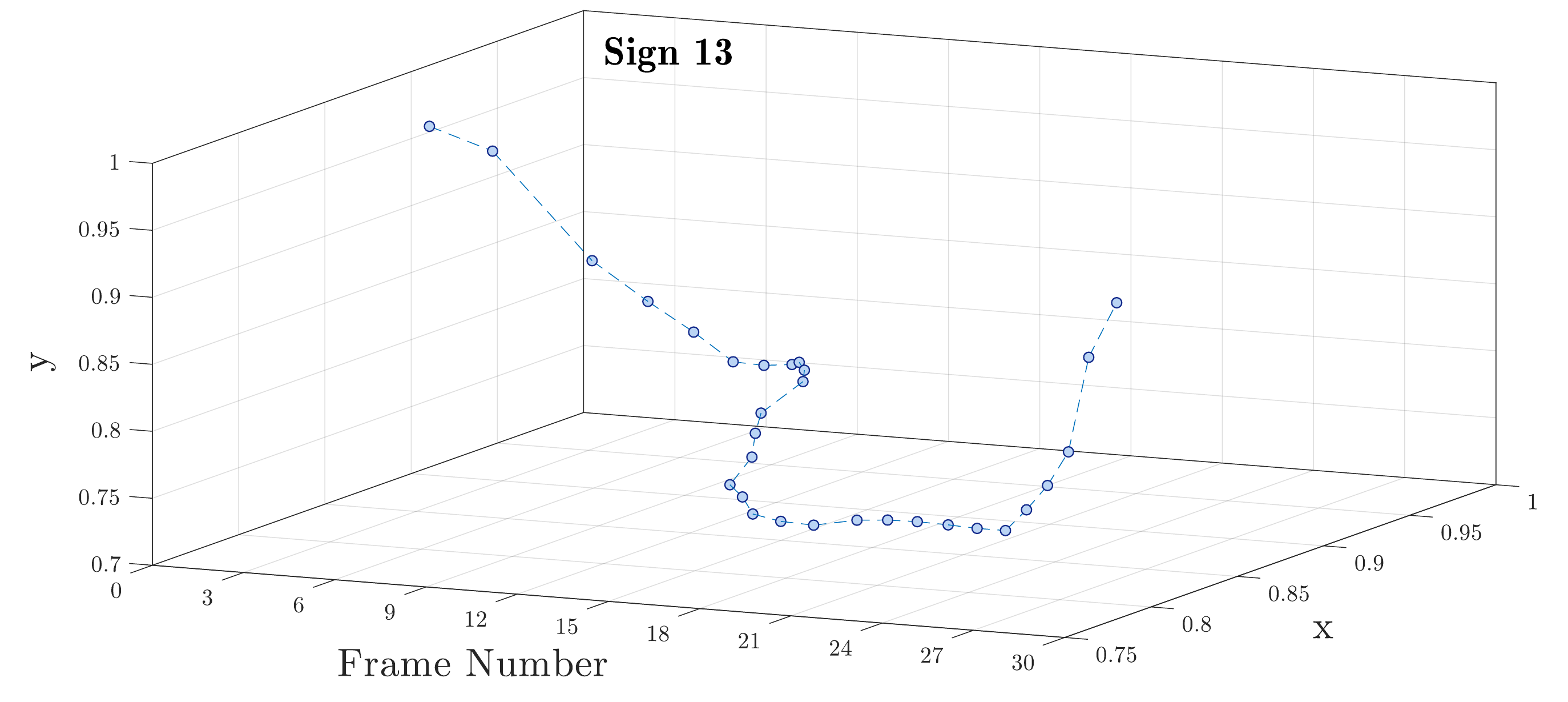}
         \caption{}
         \label{fig.fiveb}
    \end{subfigure}
\caption{Two examples of the extracted trajectories for signs 1 (sad) and 13 (mother).}
\label{fig.five}
\end{figure}

After detecting the start frame, a hand tracking process begins to extract the trajectory of the hand. This hand tracking was accomplished via a region growing technique and is illustrated in \hyperref[fig.four]{Figure \ref*{fig.four}}. Specifically, the centroid of the hand region in the start frame was obtained and was denoted as $C_{1}=(x_{1},y_{1})$. This centroid’s location was used as a seed for region growing in the frame that comes after the start frame. This region growing produces the hand region in this frame. Then, the hand region centroid of this frame denoted as $C_{2}=(x_{2},y_{2})$ was used as a seed for region growing in its next frame. \hyperref[fig.four]{Figure \ref*{fig.four}} illustrates the hand region extraction of the $(n+1)$’th frame using the hand centroid of the $n$’th frame, i.e. $C_{n}$. For all the frames of the video, this procedure was repeated, producing the hand trajectory. \hyperref[fig.five]{Figure \ref*{fig.five}} shows two examples of the extracted trajectories.
In addition to the centroid of the hand, three simple shape information were also extracted from each frame as features. These shape features include: area, orientation and eccentricity. Area determines the number of pixels in the hand region for each frame. Considering the hand region as an ellipse, orientation measures the angle between its x-axis and major axis. Eccentricity is a measure of how much the bounded ellipse of the hand deviates from being circular. These features were added to the hand centroids forming a five-dimensional feature vector for each sign.

These time-varying feature vectors will be used as the observation sequences of the HMMs. Since the signs in the dataset were performed with different subjects and each subject had his/her own speed of performing the sign, there are vast differences in the number of the frames of the videos. To decrease the subject dependency of the system, we need to normalize the number of frames before training the HMMs. For this purpose, a linear temporal interpolation with 30 query points was used to normalize the number of frames. Therefore, for each video sample we extracted a $5\times30$ feature matrix.
\section{Hidden Markov Model Based Classification}
\label{section.HMM}
Unlike static signs which create time-invariant features, dynamic signs produce features which vary in time. In order to classify these time-varying features, we need a system that can model this dynamic nature of the features. HMM has long been used for the classification of temporal patterns \cite{brand1997coupled}, and it has been proved to be successful in sign language classification \cite{starner1995visual}. This section gives a brief introduction to HMM.

HMM is a stochastic model which contains a Markov chain with an invisible or hidden sequence of states. If we denote the number of hidden states as $q$, an HMM can succinctly be represented by:
\begin{equation}\label{Eq.1}
  \lambda=(\Pi,\mathbf{A},\mathbf{B})
\end{equation}
where $\Pi$ is a $q\times1$ matrix containing the initial probabilities of the states and $\mathbf{A}$ is the transition probability matrix. $\mathbf{B}$ is called the state emission probability distribution and its components are denoted as $b_{j}$ for $j$th state. At each time, the process is in one of the hidden states and generates observations according to these emission probability distributions. The observations either can be discrete or continuous. For discrete observations, the emissions of each state are represented by probability mass functions (pmf), and for continuous observations, they are represented by probability density functions (pdf). Refer to \cite{rabiner1986introduction,rabiner1989tutorial} for detailed tutorial on HMM.

The observations used in this study are five-dimensional continuous feature vectors. Therefore, a pdf should be assigned for estimating these observations. The mixture of Gaussians is proved to be a successful method for estimating the pdf of continuous observations \cite{bashir2007object}. Hence, we model the observations of each state of the HMM with a mixture of Gaussians. Let the $d$-dimensional observation vector of each state be denoted as $\mathbf{x}$ and the state at time $t$ be denoted as $q_{t}$. The pdf of the observation at state $j$ can be modeled as:
\begin{equation}\label{Eq.2}
  b_{j}(\mathbf{x})=\mathrm{P}(\mathbf{x}|q_{t}=j)=\sum_{i=1}^{M}c_{i} \mathrm{N}(\mathbf{x},\mu_{i},\Sigma_{i})
\end{equation}
where $M$ is the number of mixing Gaussian pdfs and $c$ is the mixing parameter satisfying:
\begin{equation}\label{Eq.3}
  \sum_{i=1}^{M}c_{i}=1 .
\end{equation}
$\mathrm{N}(x,\mu,\Sigma)$ is a multivariate Gaussian distribution with corresponding mean vector $\mu$ and covariance matrix $\Sigma$.

HMM-based classification is performed in two steps, i.e. training and evaluation. In training step, an HMM is trained for each sign. That is, the parameters of the triplet $\lambda$ are estimated. This procedure is known as the \emph{training problem} of HMM. The parameters are initialized to random values and then estimated using the Baum–Welch algorithm \cite{rabiner1986introduction}. Assuming we have the total number of $S$ classes or signs, the result of training step will be $S$ trained HMMs represented as $\lambda=\{\lambda_{1},\lambda_{2}, ..., \lambda_{S}\}$. In the evaluation step, once HMMs are trained, a test sign with observation vector $\mathbf{x}$, is recognized by computing the probability of $\mathbf{x}$ given each trained HMM. This is known as the \emph{evaluation problem} of HMM and is solved using the forward-backward algorithm \cite{rabiner1986introduction}. Specifically, given a set of trained HMMs for $S$ signs, i.e. $\{\lambda_{1},\lambda_{2}, ..., \lambda_{S}\}$, and the observation sequence of the test sign, its sign label is assigned according to the following formula:
\begin{equation}\label{Eq.5}
  \underset{i=1, 2, ..., S}{\operatorname{argmax}} \mathrm{P}(x|\lambda_{i}).
\end{equation}
\section{Results and Discussion}
\label{section.result}
In this section, the performance of the proposed Persian sign language recognition system is evaluated in different experiments. Before conducting the experiments, the main parameters of the system are tuned. All the following experiments are performed with two sets of features. In the first set, which is referred to as \emph{trajectory} features, only the x-y position of the hand is used as features. In the second set, which is referred to as \emph{trajectory-shape} features, in addition to hand position, the shape information of the hand is also used as features. That is, the first set contains 2-dimensional features while the second one contains 5-dimensional features.
\subsection{Parameter Tuning}
\label{subsection.tune}
\begin{figure}
\centering
    \begin{subfigure}{0.45\textwidth}
         \centering
         \includegraphics[width=\textwidth]{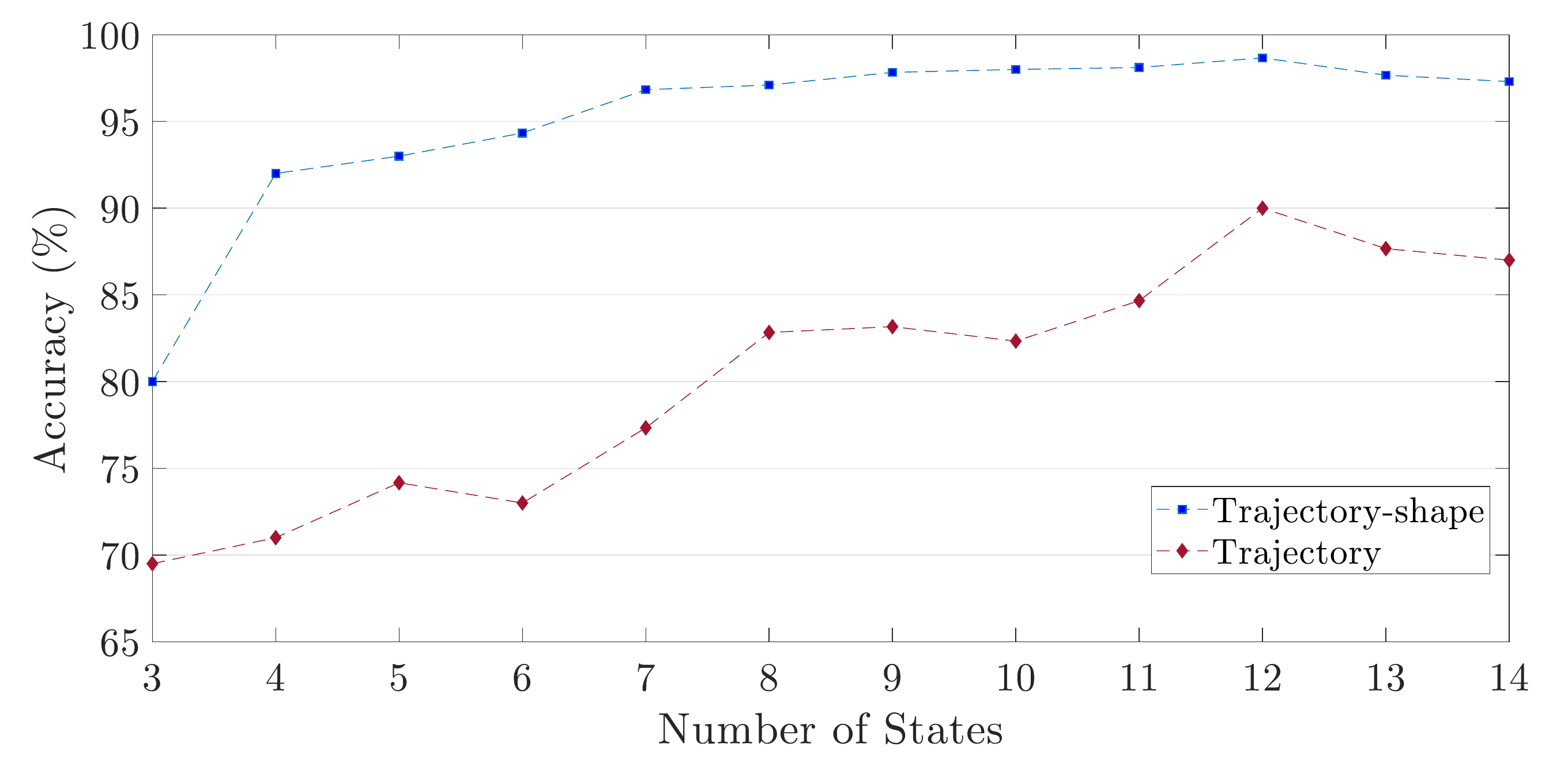}
         \caption{}
        \label{fig.sixa}
    \end{subfigure}
    \begin{subfigure}{0.45\textwidth}
         \centering
         \includegraphics[width=\textwidth]{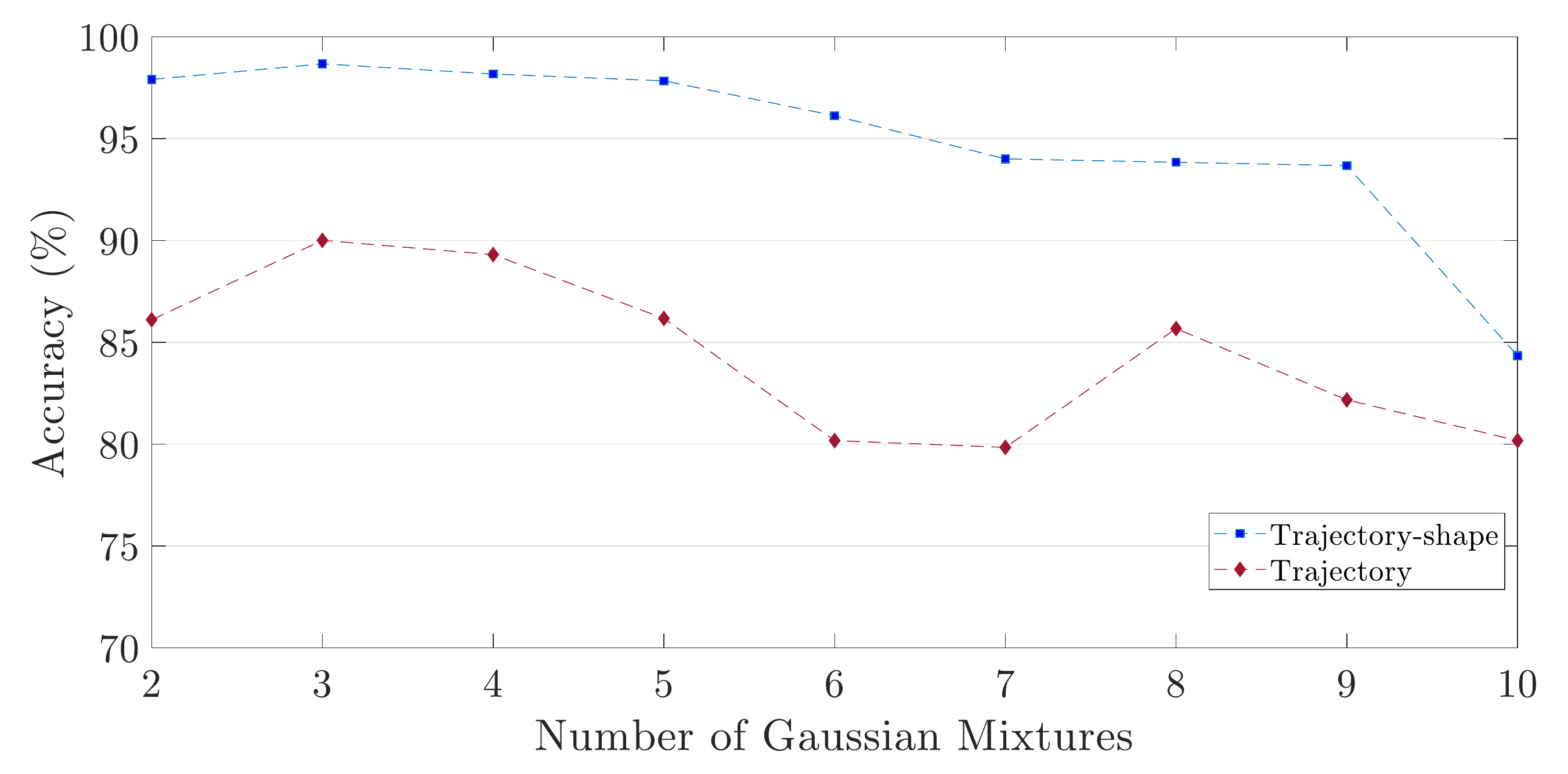}
         \caption{}
         \label{fig.sixb}
    \end{subfigure}
\caption{Tuning the number of states and the number of Gaussian mixtures. a) Classification accuracy as a function of number of states. b) Classification accuracy as a function of number of Gaussian mixtures.}
\label{fig.six}
\end{figure}

To achieve best models for sign trajectories, there are two main parameters that need to be tuned, namely the number of hidden states and number of Gaussian mixtures. For this purpose, 20\% of the samples of each sign were randomly chosen for training and the rest of the data were used for testing. For an HMM, the most important parameter is the number of hidden states. \hyperref[fig.sixa]{Figure \ref*{fig.sixa}} shows the accuracy of the system for different number of states while fixing the other parameters. It can be observed that for both set of features, the accuracy of the system increases as we increase the number of states from 3 to 12. The highest accuracy of the system is achieved for 12 states for both sets of features, i.e., 98.66\% for trajectory-shape and 91.2\% for trajectory features.

\begin{figure}
\centering
    \begin{subfigure}{0.45\textwidth}
         \centering
         \includegraphics[width=\textwidth]{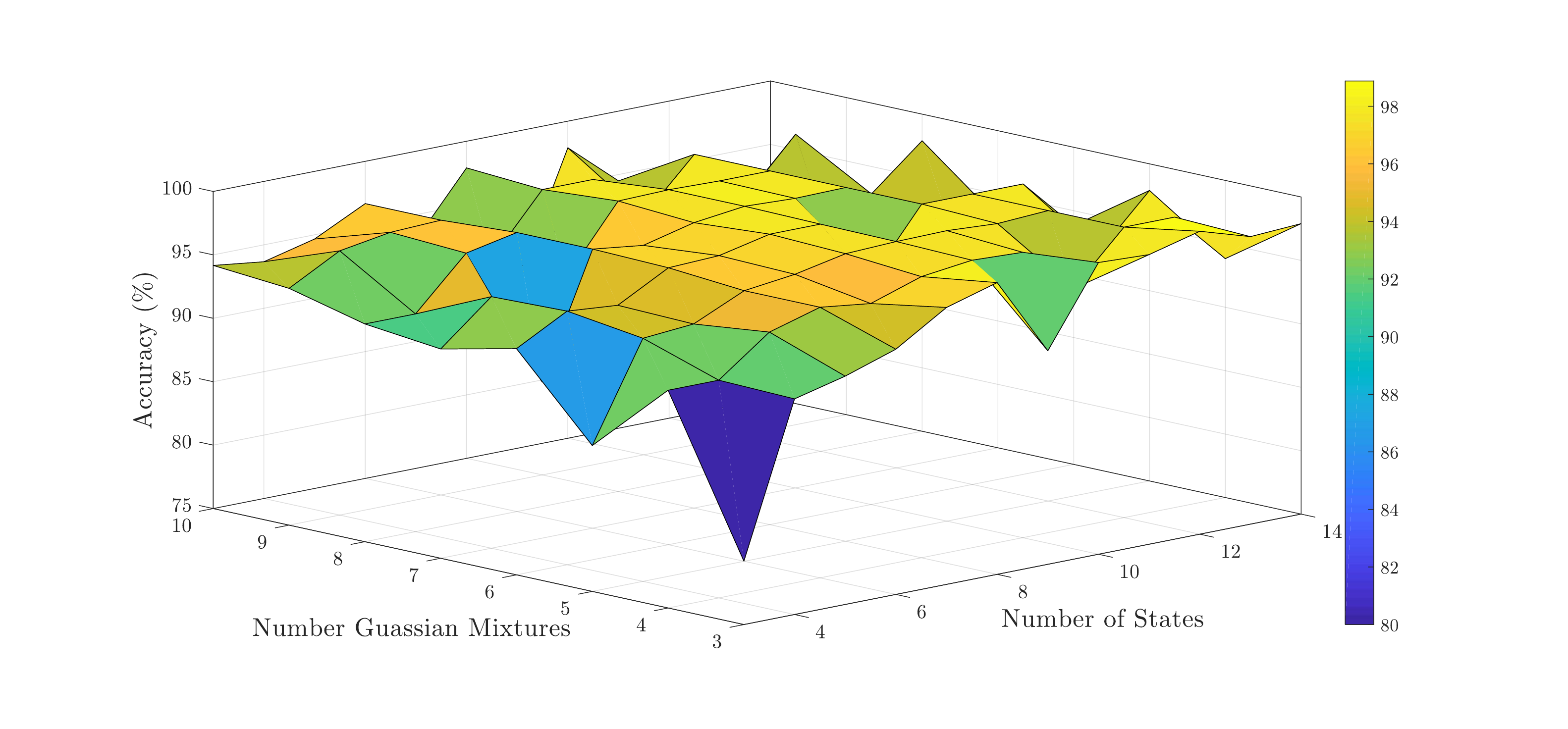}
         \caption{}
        \label{fig.sevena}
    \end{subfigure}
    \begin{subfigure}{0.45\textwidth}
         \centering
         \includegraphics[width=\textwidth]{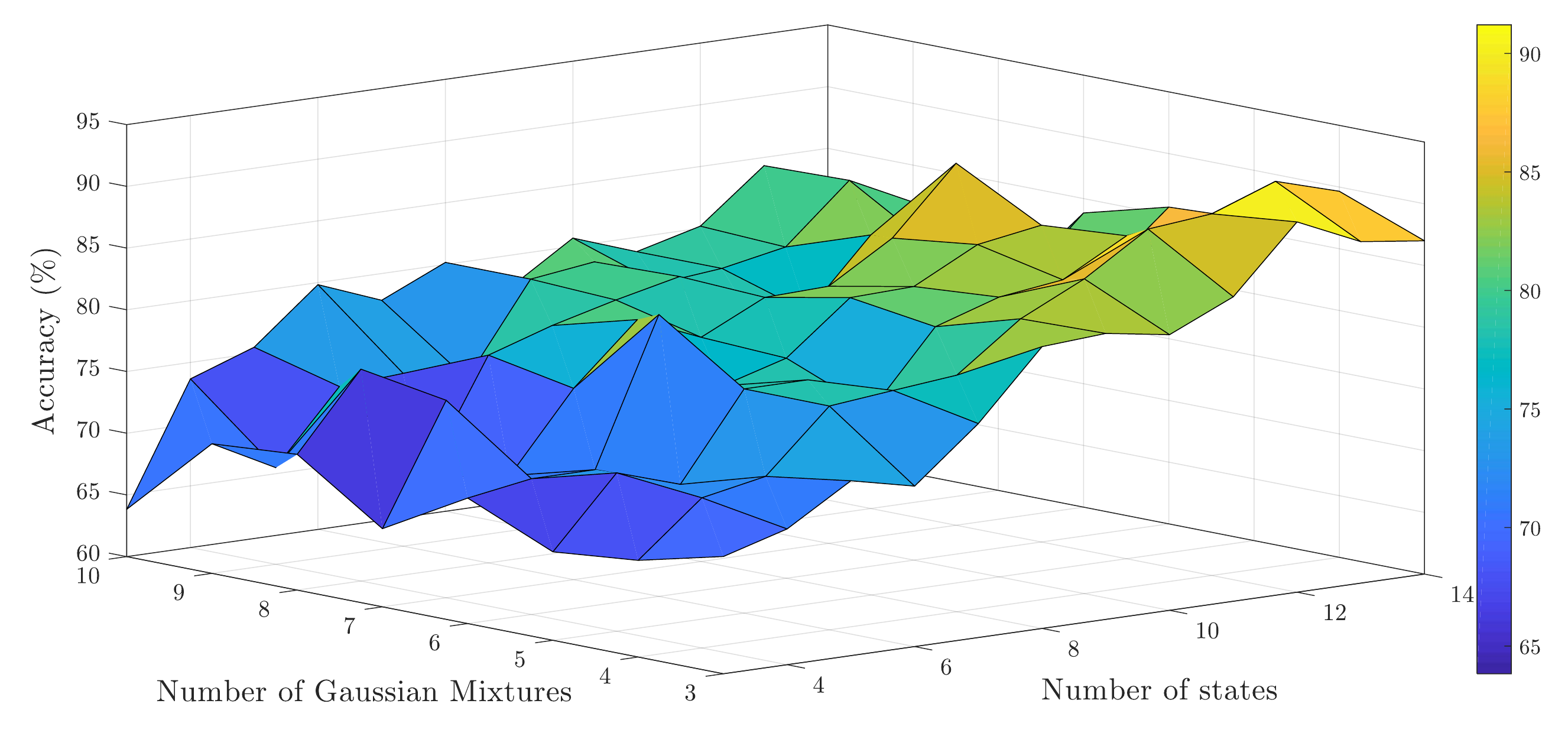}
         \caption{}
         \label{fig.sevenb}
    \end{subfigure}
\caption{Classification accuracy as a function of the number of states and number of Gaussian mixtures. a) For trajectory-shape features. b) For Trajectory features.}
\label{fig.seven}
\end{figure}
The next parameter to be tuned is the number of Gaussian mixtures. \hyperref[fig.sixb]{Figure \ref*{fig.sixb}} shows the accuracy of the system for different number of Gaussian mixtures while fixing the other parameters. For both sets of features, the best accuracy is achieved for 3 mixtures, and it decreases as we increase the number of mixtures, revealing that 3 mixture of Gaussians is the best representation for our observation data. To more evaluate the role of these two parameters, \hyperref[fig.seven]{Figure \ref*{fig.seven}} presents the classification performance for varying number of states and Gaussian mixtures. From this figures similar deduction to \hyperref[fig.six]{Figure \ref*{fig.six}} can be made about the optimal number of states and mixtures. Moreover, it can be observed from the figure that the trajectory-shape set of features (\hyperref[fig.sevena]{Figure \ref*{fig.sevena}}) is more robust to these parameters than the trajectory features (\hyperref[fig.sevenb]{Figure \ref*{fig.sevenb}}). To summarize, considering the results presented in \hyperref[fig.six]{Figure \ref*{fig.six}} and \hyperref[fig.seven]{Figure \ref*{fig.seven}}, the optimal number of states and Gaussian mixtures were set to 12 and 3, respectively.
\subsection{Sign Classification}
\label{subsection.classification}
In this section, the classification results of 20 dynamic Persian signs are presented. After extracting the hand trajectory and shape information, 20 HMMs were trained using both sets of features with 12 hidden states and 3 mixtures of Gaussian. In addition to HMM, Support Vector Machine (SVM) with polynomial kernel was also used for classification of signs and the results were compared to the ones obtained from HMM classification.

Three different training strategies with 20\% of the samples of each sign were used to evaluate the performance of the system, namely random, subject-dependent and subject-independent training. In random training strategy, as its name suggests, 20\% of the samples were selected randomly as training data, leaving the rest of the data for testing. In signer-dependent strategy, 20\% of the samples of each subject were selected as training data. Considering we have five samples from each subject, one sample per subject was selected as training data. That is, we made sure that each subject had a sample among training data. In subject-independent strategy, on the other hand, we trained the system with samples of only two subjects, and the samples from the other ten subjects were left for testing. The results are presented in \hyperref[table.result]{Table \ref*{table.result}}. All the experiments were conducted in 10 runs, and the mean and variance values of the classification accuracy are reported in \hyperref[table.result]{Table \ref*{table.result}}.
\begin{table}
  \centering
\caption{Classification results for different classifiers with 20\% of the samples used for training in different training strategy.}
\label{table.result}
\begin{tabular}{c|c c|c c}
  \hline
  Classifier & \multicolumn {2}{c|}{SVM} & \multicolumn {2}{c}{HMM} \\
  Feature set & Trajectory & Trajectory-shape & Trajectory & Trajectory-shape \\
  \hline \hline
  Random & 78.47 ($\pm$0.85) & 87.77 ($\pm$1.12) & 87.12 ($\pm$0.21) & \textbf{98.13 ($\pm$0.11)} \\
  Subject-dependent & 83.54 ($\pm$0.62) & 89.79 ($\pm$0.23) & 87.01 ($\pm$0.13) & \textbf{97.63 ($\pm$0.12)} \\
  Subject-independent & 62.90 ($\pm$0.75) & 67.10 ($\pm$0.41) & 83.20 ($\pm$0.31) & \textbf{96.70 ($\pm$0.09)} \\
  \hline
\end{tabular}
\end{table}

Some observations can be made from this table. First, in both classifiers, adding shape information to the trajectory features has significantly (10\%) increased the accuracy of the system, indicating the importance of the hand shape information in sign classification. Second, the results obtained by HMM is notably better than SVM. This may be due to the inability of SVM in dealing with the time-varying nature of the features, while HMM can successfully model these temporal features. Third, considering different training strategies, the results obtained by SVM meaningfully decrease in the subject-independent case and slightly increase for subject-dependent training. This leads to the conclusion that the system designed with SVM as a classifier is extremely subject-dependent. Contrary to SVM, HMM represents excellent subject-independent results and the classification accuracy drops only 1.4\% in subject-independent case.

\begin{figure}
\centering
    \begin{subfigure}{0.45\textwidth}
         \centering
         \includegraphics[width=\textwidth]{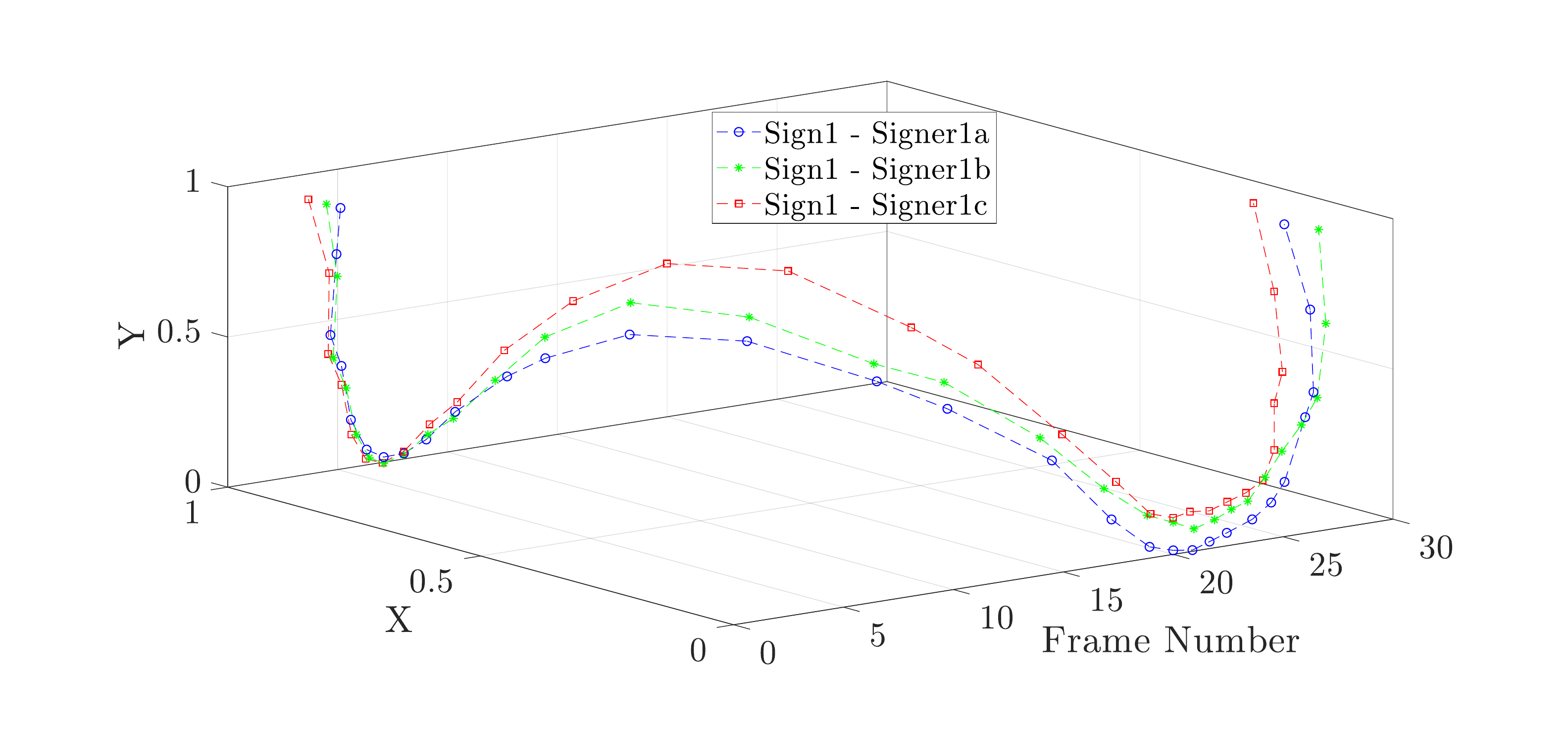}
         \caption{}
        \label{fig.eighta}
    \end{subfigure}
    \begin{subfigure}{0.45\textwidth}
         \centering
         \includegraphics[width=\textwidth]{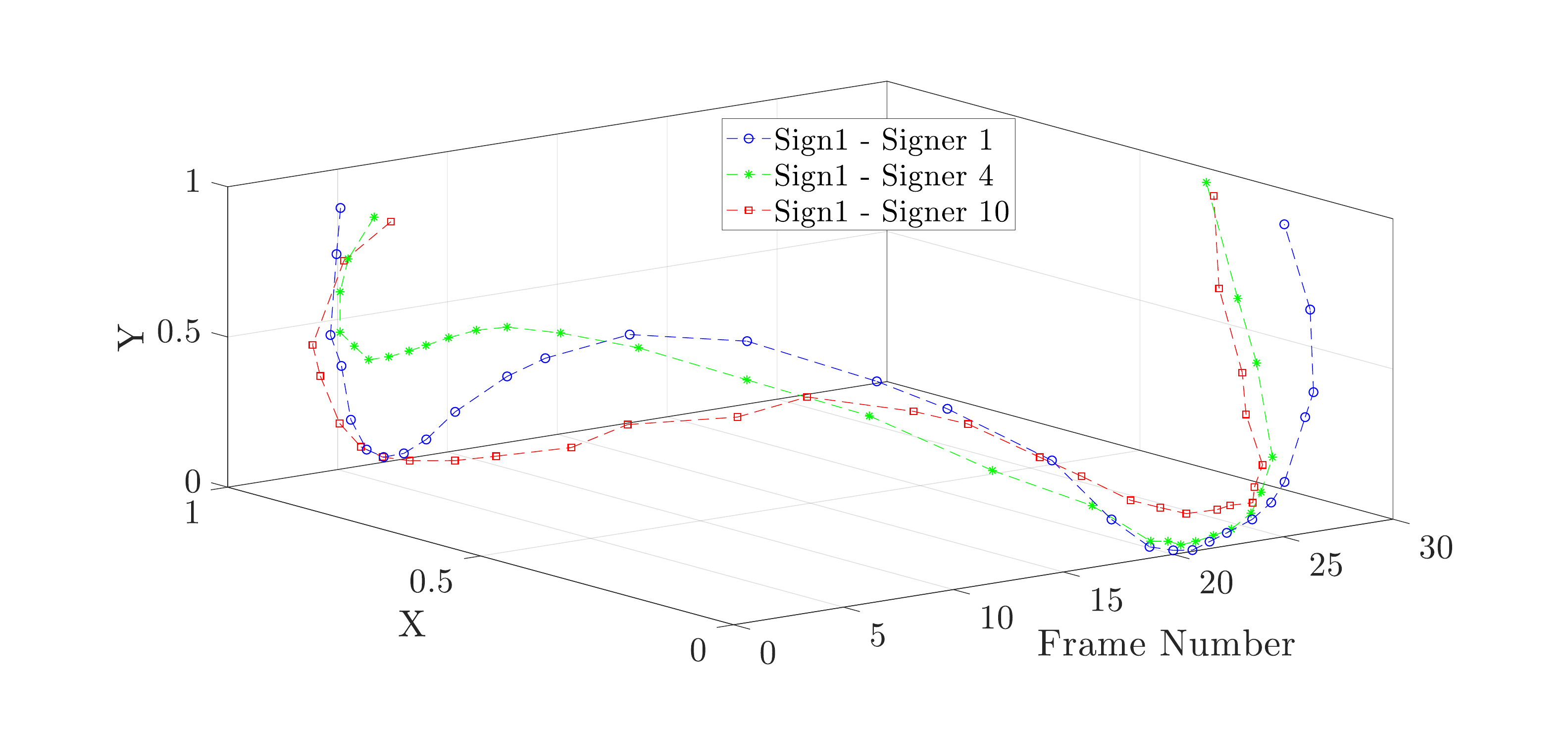}
         \caption{}
         \label{fig.eightb}
    \end{subfigure}
\caption{Samples of the same sign performed by (a) a single signer and (b) three different signers.}
\label{fig.eight}
\end{figure}

To further discuss the concept of signer-independency, \hyperref[fig.eight]{Figure \ref*{fig.eight}} presents samples of the same sign performed by a single signer (\hyperref[fig.eighta]{Figure \ref*{fig.eighta}}) and three different signers (\hyperref[fig.eightb]{Figure \ref*{fig.eightb}}). As it can be seen in \hyperref[fig.eighta]{Figure \ref*{fig.eighta}}, different realizations of a sign performed by a single signer are very similar in terms of both the shape of the trajectory and the x and y values of each frame number. For different signers (\hyperref[fig.eightb]{Figure \ref*{fig.eightb}}), although the x and y values of each frame are different, the shape of the trajectory is almost similar. SVM treats each frame as a separate feature and fails to see the temporal pattern of the features. As a result, it hardly recognizes the similarities between the signs performed by different signers. Therefore, exhibits weak signer-independent results. HMM, on the other hand, can model this temporal patterns using its state transition capabilities.
\begin{figure}
         \centering
         \includegraphics[width=\textwidth]{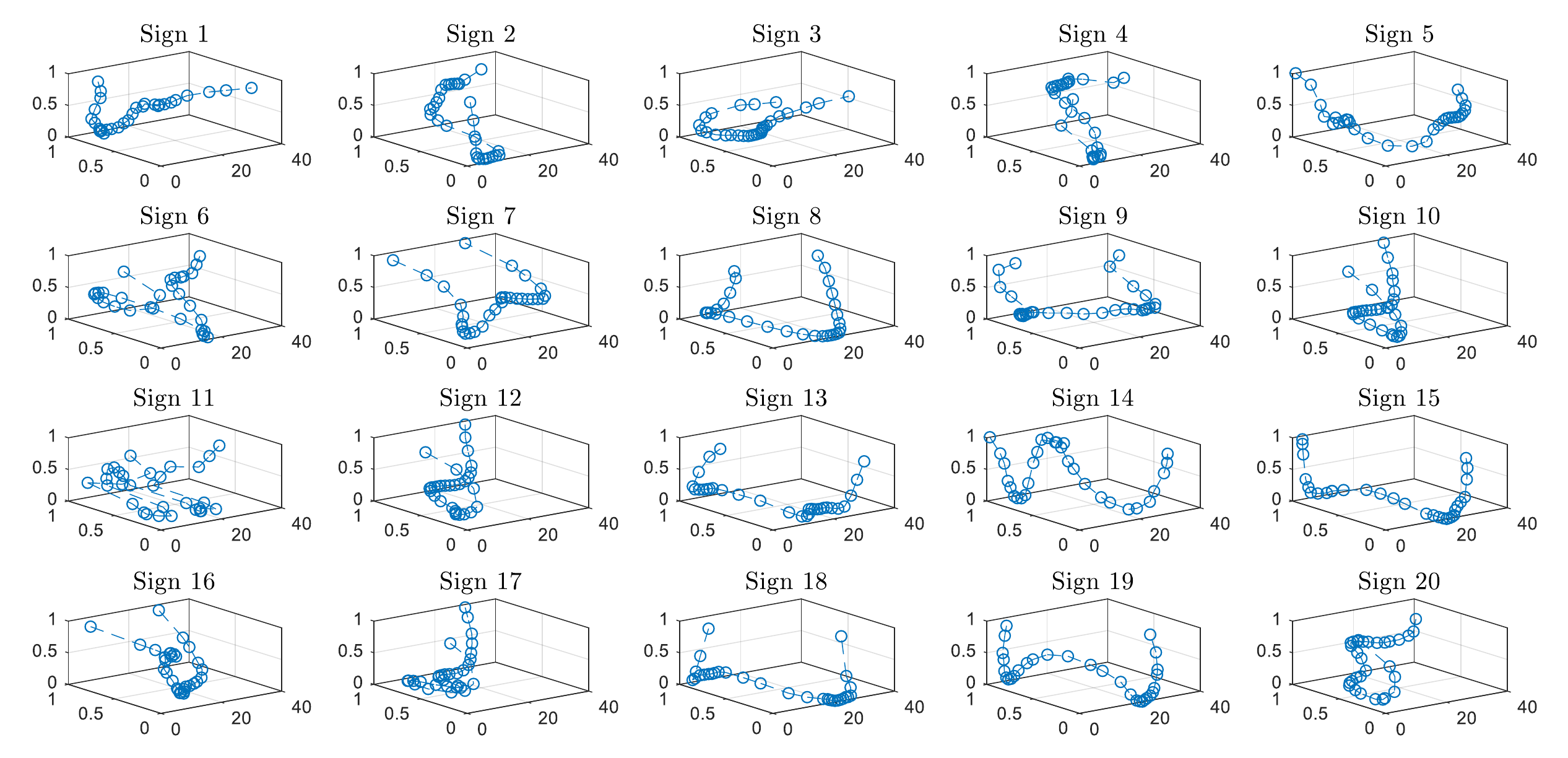}
         \caption{The trajectories of the 20 persian signs used in this study. Note that the labels are eliminated to decrease the ambiguity of the figure.}
         \label{fig.nine}
\end{figure}
\begin{figure}
\centering
    \begin{subfigure}{0.3\textwidth}
         \centering
         \includegraphics[width=\textwidth]{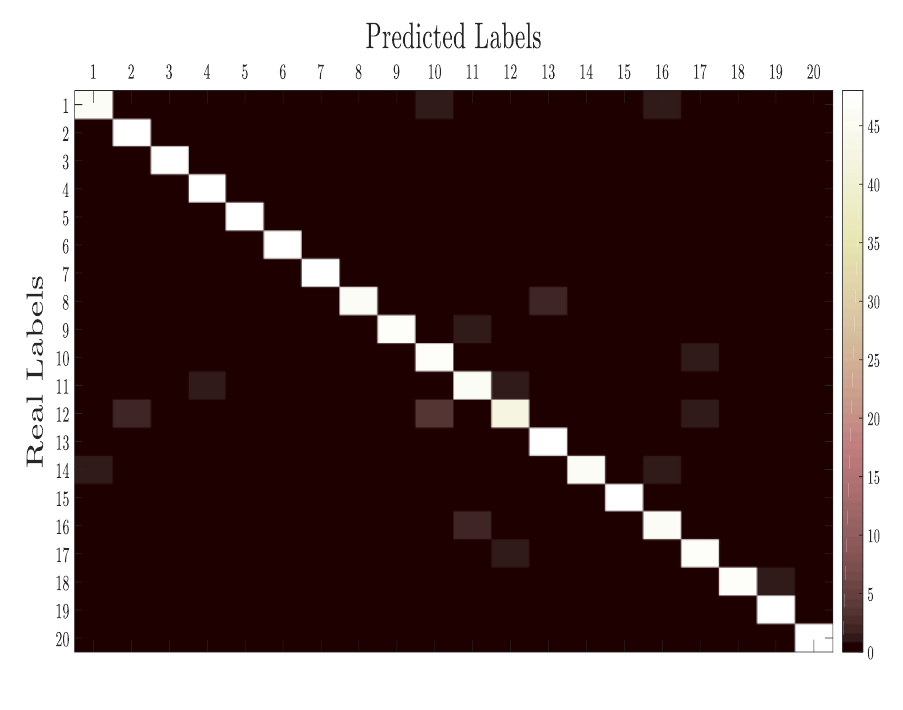}
         \caption{}
        \label{fig.tena}
    \end{subfigure}
    \begin{subfigure}{0.45\textwidth}
         \centering
         \includegraphics[width=\textwidth]{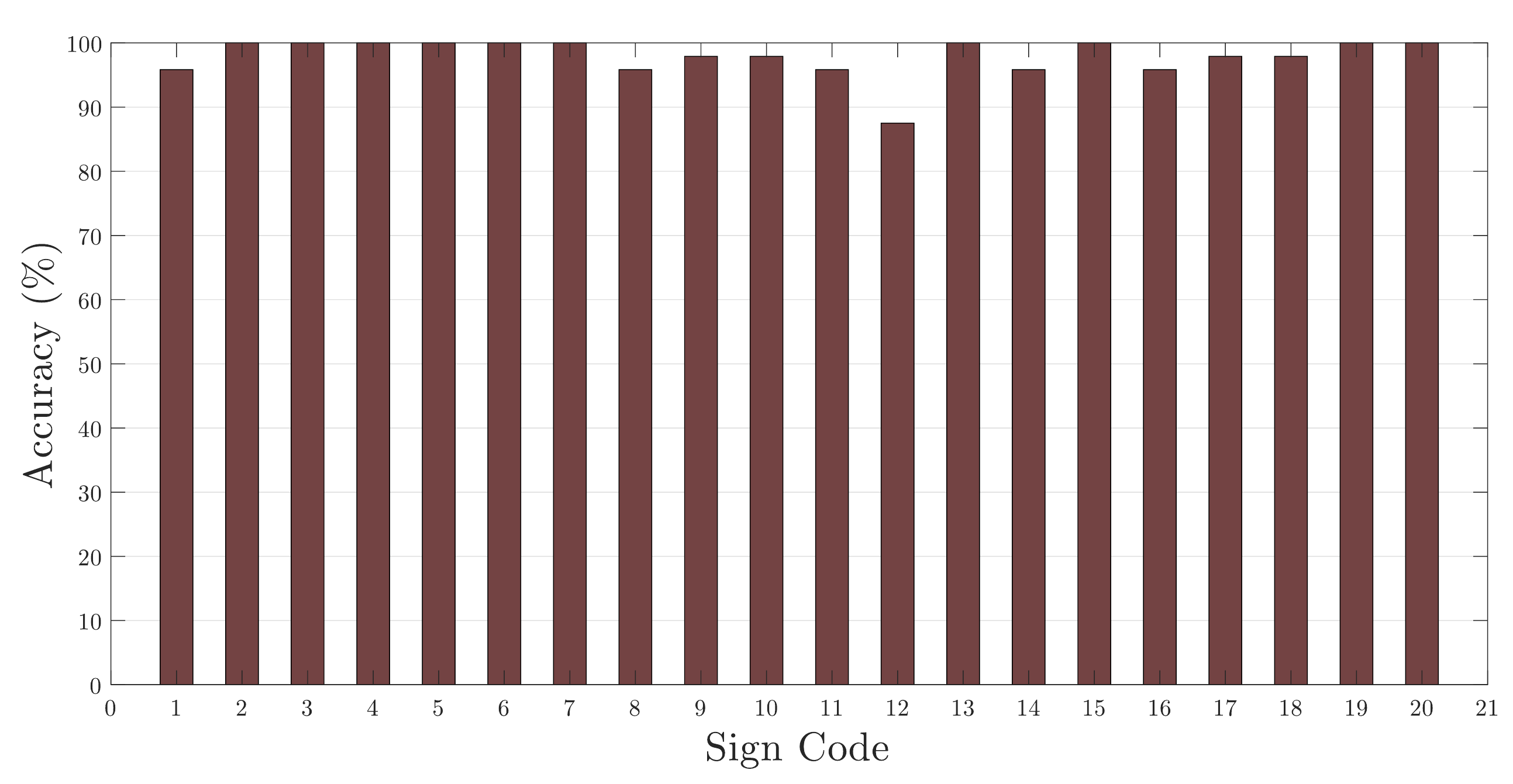}
         \caption{}
         \label{fig.tenb}
    \end{subfigure}
\caption{Performance of the proposed HMM-based system. Each sign is represented by its corresponding code. a) Confusion matrix of the classification. b) Accuracies obtained for each class of signs.}
\label{fig.ten}
\end{figure}

The trajectories of the 20 signs of the dataset are illustrated in \hyperref[fig.nine]{Figure \ref*{fig.nine}}. This figure portrays the spread of realizations between these 20 signs. As it can be seen in this figure, most of the signs are distinguishable while a few of them have similar trajectories that can challenge the performance of the system. To better evaluate the performance of the proposed HMM-based system, the confusion matrix of the classification and the accuracies obtained for each class of signs are illustrated in \hyperref[fig.tena]{Figure \ref*{fig.tena}} and \hyperref[fig.tenb]{Figure \ref*{fig.tenb}}, respectively. In these figures each sign is represented by its corresponding code from \hyperref[table.list]{Table \ref*{table.list}}. According to these figures and the sign trajectories of \hyperref[fig.nine]{Figure \ref*{fig.nine}}, following observations can be made. Half of the signs (signs 2-7, 13, 15, 19 and 20) are classified with 100\% accuracy. Among these signs, signs 2-7 and sign 20 have distinguishable trajectories and the obtained accuracy was predictable. Signs 13, 15 and 19 have similar trajectories but they have been classified with 100\% accuracy. This can be explained by the added shape information that has enable the system to discriminate between these signs.  The lowest accuracy is obtained for sign 12 and it is mainly misclassified with sign 10, which may be due to their similar trajectories.

\begin{figure}
         \centering
         \includegraphics[width=0.7\textwidth]{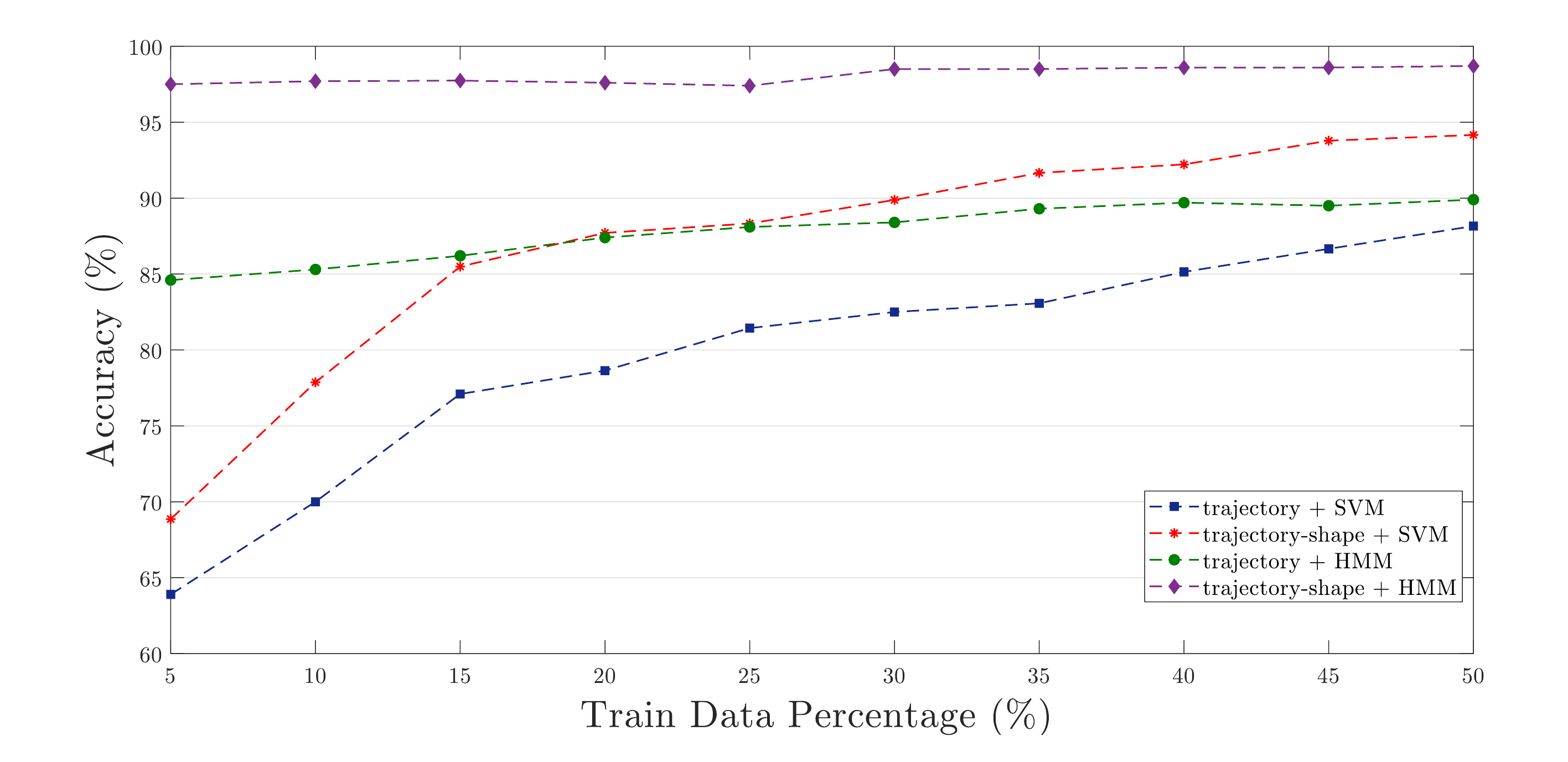}
         \caption{The accuracy of the examined methods as a function of the train data percentage.}
         \label{fig.eleven}
\end{figure}
One of the most critical aspects of a recognition system is its level of dependence on the number of training data. Regarding the challenges in the acquisition of the sign videos, the number of available samples for each class is usually restricted. Therefore, it is essential for a recognition system to perform correctly with limited training data. \hyperref[fig.eleven]{Figure \ref*{fig.eleven}} exhibits the accuracy of the examined methods as a function of the train data percentage. It can be seen that the SVM-based methods rely significantly on the number of training data and their performance decrease as we decrease the train data, whereas HMM-based methods, especially the method with trajectory-shape features, are entirely robust to the number of train data. It can be observed from the figure that even with 5\% of the data for training, the system can successfully model the signs, and for train data percentage of more than 30, the performance of the system remains almost the same. Note that the accuracy is not yet at ceiling for training data percentage of 50.

\section{Conclusion}
\label{section.conclusion}
In this study, a dynamic Persian sign language recognition system is presented. A dataset containing 1200 videos of 20 signs were collected. Hand trajectories along with three hand shape information were extracted from video frames using a region growing technique. HMM with Gaussian mixture observations was utilized to model these trajectories and their temporal patterns. According to the experimental results, the HMM-based system with hand’s trajectory and shape information as features can successfully recognize these 20 signs with an average accuracy of 98.13\%. Moreover, the experiments indicated that the performance of the system is independent of the subject, and it has excellent performance even with a limited number of training data.

This study being an initial study on dynamic PSL recognition has only focused on the trajectories of the signs. While, it is likely that using a wider dictionary of signs will increase the possibility of more similar trajectories leading to a need for more training data. This problem can be addressed by using two cameras to extract both spatial and depth information and decrease the possibility of similar trajectories. Another solution may be to use more sophisticated approaches like deep learning based features. For future studies, the authors will be focused on updating the dataset and using deep learning based approaches for PSL recognition.
\section*{Acknowledgment}
This paper is published as part of a research project supported by the University of Tabriz, Research Affairs Office, Iran. The authors would like to thank the Society of Deaf People (SDP), Urmia, Iran, for the many valuable assistances they provided during the acquisition of the dataset.

%\mathbf{Y}_{t}-\mathbf{D}_{t-1}\mathbf{X}_{t}\rVert_{\mathrm{F}}^{2}+\mu_{1}\lVert \mathbf{X}_{t}\rVert_{2,1}+\mu_{2}\mathrm{\psi}(\mathbf{X}_{t})\right\}
\bibliographystyle{elsarticle-num}
\bibliography{references}

\end{document}